\pdfoutput=1

\documentclass[11pt]{article}

\usepackage[]{ACL2023}

\usepackage{times}
\usepackage{latexsym}

\usepackage[T1]{fontenc}

\usepackage[utf8]{inputenc}

\usepackage{microtype}

\usepackage{inconsolata}

\usepackage{markdown}

\usepackage{multirow}
\usepackage{booktabs}
\usepackage{amsmath}

\usepackage{titlesec}
\titleformat{\subsection}[runin]{\normalfont\bfseries}{}{0em}{}[.\;\;\;]
\titlespacing*{\subsection}{0pt}{0.7em}{0pt}

\usepackage[color=orange!50!white]{todonotes}
\usepackage[nameinlink]{cleveref}
\usepackage{placeins}

\usepackage{floatrow}
\usepackage{tabularx}
\usepackage{csquotes}
\usepackage{amssymb}

\usepackage{xcolor}
\definecolor{anne}{RGB}{153,205,255}

\title{BoschAI @ PLABA 2023: Leveraging Edit Operations in End-to-End Neural Sentence Simplification}

\author{
    \bf Valentin Knappich$^{1,2}$ \hspace{6mm} Simon Razniewski$^1$ \hspace{6mm} Annemarie Friedrich$^2$ \\ \\
    $^1$Bosch Center for Artificial Intelligence, Renningen, Germany \hspace{2mm}$^2$University of Augsburg, Germany\\
    \texttt{valentin.knappich|simon.razniewski@de.bosch.com} \\
    \texttt{annemarie.friedrich@informatik.uni-augsburg.de}
}

\begin{document}

\maketitle

\begin{abstract}
    Automatic simplification can help laypeople to comprehend complex scientific text.
Language models are frequently applied to this task by translating from complex to simple language.
In this paper, we describe our system based on Llama 2, which ranked first in the PLABA shared task addressing the simplification of biomedical text. 
We find that the large portion of shared tokens between input and output leads to weak training signals and conservatively editing models.
To mitigate these issues, we propose sentence-level and token-level loss weights.
They give higher weight to modified tokens, indicated by edit distance and edit operations, respectively.
We conduct an empirical evaluation on the PLABA dataset and find that both approaches lead to simplifications closer to those created by human annotators ($+1.8\%$ / $+3.5\%$ SARI), simpler language ($-1$ / $-1.1$ FKGL) and more edits ($1.6x$ / $1.8x$ edit distance) compared to the same model fine-tuned with standard cross entropy.
We furthermore show that the hyperparameter $\lambda$ in token-level loss weights can be used to control the edit distance and the simplicity level (FKGL).
\end{abstract}

\section{Introduction}\label{sec:intro}
\begin{figure}[t]
    \centering
    \includegraphics[width=\linewidth]{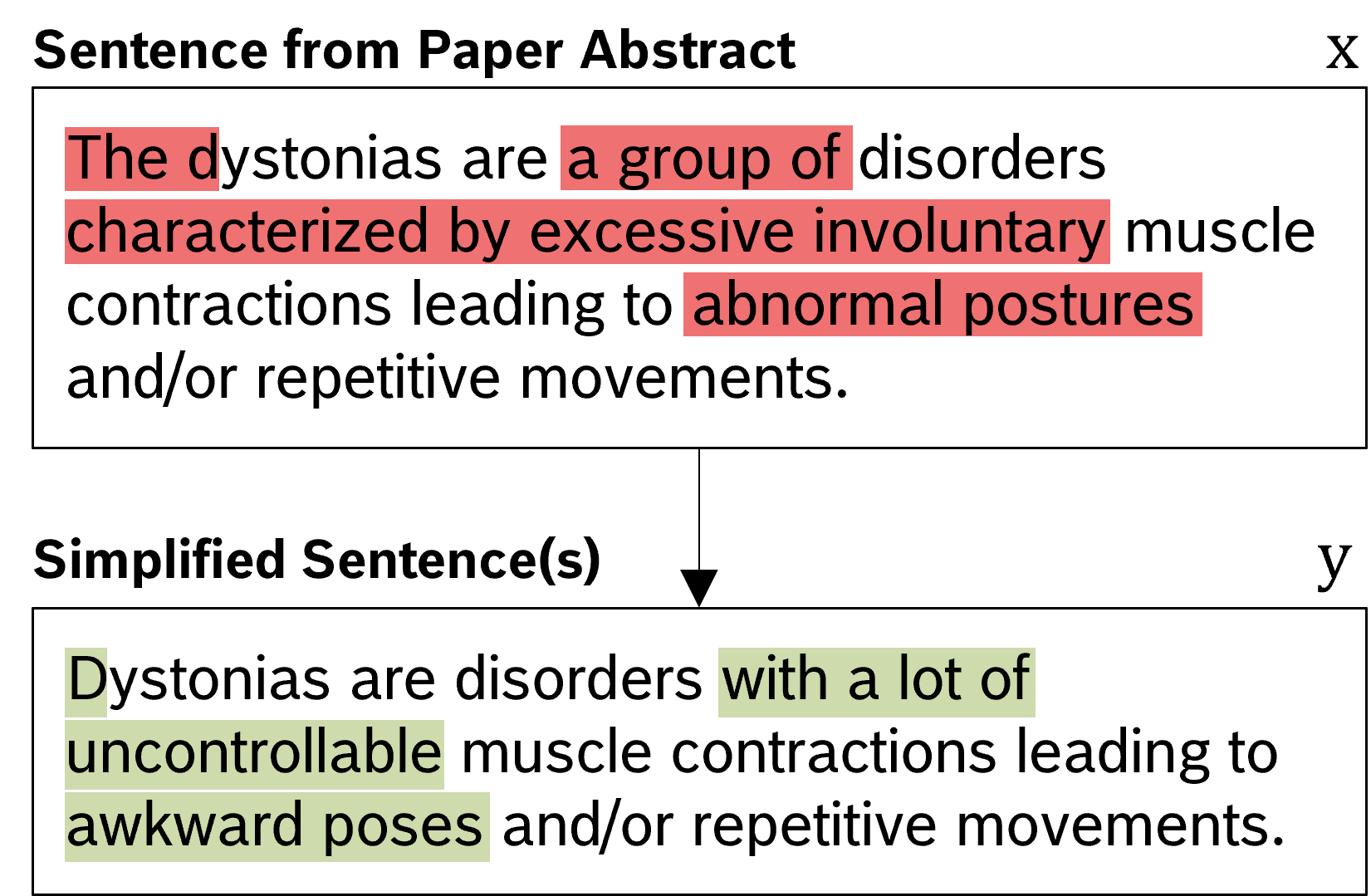}
    \caption{Example sentence pair from the PLABA dataset. Red background indicates source tokens that were removed during simplification, green background indicates target tokens that were added, white background indicates tokens that remained unchanged.}
    \label{fig:example}
\end{figure}

Scientific texts are increasingly available on the internet, and constitute a giant collection of knowledge \cite{ronzanoKnowledgeExtractionModeling2016}. However, they are also difficult to comprehend for laypeople, due to complex sentence structure and terminology \cite{gopenScienceScientificWriting1990}. Automatic text simplification systems alleviate this issue by transforming complex text into easier-to-understand simplifications.
Such simplifications ideally contain the same essential information, but convey it in a simpler, more comprehensible fashion \cite{ondovSurveyAutomatedMethods2022}.

Text simplification is especially important in scientific domains that are directly relevant to the general public, like the biomedical sciences. To facilitate work on simplification of biomedical text, the PLABA dataset was introduced by \citet{attalDatasetPlainLanguage2023}.
It contains sentence-aligned simplifications for the abstracts of 750 biomedical papers.
This paper reports our submission to the PLABA 2023 shared task\footnote{\url{https://bionlp.nlm.nih.gov/plaba2023/}} organized by NIST\footnote{\url{https://www.nist.gov/}}, the results of which are presented in \autoref{tab:challenge_results}.

\begin{table}
    \begin{tabular}{lccc}
        \toprule
        Submission  & SARI           & Simp. avg      & Acc. avg       \\
        \midrule
        Bee\_Man    & 40.59          & 82.74          & 75.00          \\
        MasonNLP    & 39.51          & 91.28          & 92.52          \\
        PLABA\_base & 35.97          & \textbf{93.53} & \textbf{94.87} \\
        \midrule
        Ours        & \textbf{44.65} & 88.86          & \textbf{94.87} \\
        \bottomrule
    \end{tabular}
    \caption{Shared Task Results on the blind test set. SARI is automatically evaluated against 4 references. Simp. avg and Acc. avg are average human evaluation scores across multiple criteria concerning simplicity and accuracy, repsectively.}
    \label{tab:challenge_results}
\end{table}

Various approaches have been applied to the task of text simplification, including lexical, syntactic, statistical and neural approaches \cite{ondovSurveyAutomatedMethods2022}.
Statistical and neural approaches essentially treat the task as a translation task, translating from complex to simple language.
\autoref{fig:example} shows an example sentence pair, where the edit operations performed by the human annotator are indicated using red and green colors.
In contrast to cross-lingual translations, simplifications share a significant portion of the tokens with the input.
We argue that this leads to suboptimal training because the gradients from most tokens point to the trivial solution of copying that token.
We indeed find that models fine-tuned on such data tend to edit much less than human annotators.
To incentivize the model to edit more, we lend the concept of edit operations (keep, replace, insert, delete), which are also used in edit-based simplification \cite{dongEditNTSNeuralProgrammerInterpreter2019,kumarIterativeEditBasedUnsupervised2020,xuEDITOREditBasedTransformer2021}. However, rather than directly predicting the edit operations, we leverage them to inform the training process and correct the ill-posed objective created by the shared tokens. In particular, we use the edit operations to derive loss weights that assign higher importance to edited tokens (token-level loss weights) or to highly edited sentences (sentence-level). Our core contributions are:

\begin{enumerate}
    \item We propose and evaluate sentence-level and token-level loss weights for text simplification and find that both are effective in increasing the edit distance of the generated outputs, while keeping the outputs faithful.
    \item We investigate the relationship between edit distance and simplification similarity measured by SARI \cite{xuOptimizingStatisticalMachine2016}, and find that both loss weighting schemes also significantly improve SARI scores.
    \item We perform a qualitative analysis of generated outputs to provide further insights beyond the quantitative results.
\end{enumerate}

\section{Related Work}\label{sec:relwork}
In this section, we present the most relevant previous works from the fields of biomedical text simplification, controllable simplification and weighted loss functions.

\subsection{PLABA}

\textit{Plain Language Adaptation of Biomedical Abstracts} (PLABA, \citealp{attalDatasetPlainLanguage2023}) is a dataset containing 750 biomedical abstracts and 921 corresponding simplifications. The dataset was created by scraping 75 common medical questions from a forum, and retrieving the abstracts of 10 relevant papers per question from PubMed. 
Human annotators were instructed in the guidelines to replace arcane words with common synonyms, split complex sentences and omit sentences that are irrelevant to consumer understanding.
All simplified abstracts are sentence-aligned, i.e., for every source sentence, there are zero, one, or multiple target sentences.

\subsection{Biomedical Text Simplification}

\citet{ondovSurveyAutomatedMethods2022} survey models, evaluation methods and datasets for biomedical text simplification. They identify data availability as one key challenge in the application of neural models to the task.
\citet{guoAutomatedLayLanguage2022} investigate the usage of language models to improve health literacy. They scrape a dataset of biomedical abstracts and corresponding summarizations, and train models to simplify and summarize the abstracts. Their models simplify the language while additionally compressing the information. In contrast, PLABA \citep{attalDatasetPlainLanguage2023} focuses on the simplification aspect of the task, where the average text length even increases slightly.

\subsection{Controllable Text Simplification}

Controllable text simplification aims to give users of language models control over the target complexity of generated outputs. To that end, some works \cite{martinControllableSentenceSimplification2020,liInvestigationEffectControl2022} employ learned control tokens that condition the language model to generate text in the desired complexity. Others use reinforcement learning \cite{yanamotoControllableTextSimplification2022} or lexically constrained decoding \cite{zetsuLexicallyConstrainedDecoding2022}. In this work, we also build on edit operations, but use them to improve the training process of autoregressive language models rather than predicting them directly.

\subsection{Weighted Loss Functions}

In weighted loss functions, losses from individual samples are weighted differently. This is commonly used to counteract issues arising from imbalanced training sets \cite{henningSurveyMethodsAddressing2023}. In the context of text simplification, complexity-based loss weights have been used to incentivize the model to generate simpler text \cite{krizComplexityWeightedLossDiverse2019}. The authors derived a degree of complexity for every word in the vocabulary by analyzing their occurrences in texts of different complexity levels. These complexity levels were then used as loss weights, such that simpler words have a higher weight. In the empirical evaluation, this modified loss function led to more insertions, but also a higher FKGL and longer sentences.

\begin{table*}[t]
    \resizebox{\textwidth}{!}{%
        \begin{tabular}{llccccccccc}
            \toprule
                                                                         &                        & \multicolumn{3}{c}{Similarity} &                 & \multicolumn{3}{c}{Simplicity} &  & \multirow{2}{*}{Edit distance}                                                           \\
            \cmidrule{3-5} \cmidrule{7-9}
                                                                         &                        & SARI                           & BLEU            & ROUGE-L                        &  & FKGL                           & Difficult Words & \#words         &  &                  \\
            \midrule
            \multirow{2}{*}{\rotatebox[origin=c]{90}{\small Baselines}}  & Copy                   & $15.6$                         & $27.0$          & $53.9$                         &  & $13.8$                         & $8.5$           & $26.2$          &  & $0.0$            \\
                                                                         & Empty                  & $21.7$                         & $0.7$           & $0.0$                          &  & $-15.7$                        & $0.0$           & $0.0$           &  & $154.7$          \\
                                                                         & Ground Truth           & $81.5$                         & $98.8$          & $98.1$                         &  & $11.3$                         & $7.3$           & $27.6$          &  & $89.7$           \\
            \cmidrule{1-11}
            \multirow{2}{*}{\rotatebox[origin=c]{90}{\small Zero-Shot}}  & Llama-2-7B             & $29.8 \pm 0.07$                & $6.4 \pm 0.10$  & $20.5 \pm 0.03$                &  & $8.7 \pm 0.06$                 & $12.1 \pm 0.17$ & $89.8 \pm 0.13$ &  & $355.3 \pm 0.40$ \\
                                                                         & Llama-2-13B            & $28.1 \pm 0.41$                & $6.8 \pm 0.07$  & $20.5 \pm 0.12$                &  & $8.2 \pm 0.19$                 & $12.6 \pm 0.11$ & $84.7 \pm 0.52$ &  & $331.6 \pm 2.53$ \\
                                                                         & ChatGPT                & $33.0 \pm 0.10$                & $7.0 \pm 0.09$  & $31.3 \pm 0.14$                &  & $10.5 \pm 0.34$                & $6.6 \pm 0.23$  & $31.8 \pm 0.11$ &  & $118.3 \pm 0.63$ \\
            \cmidrule{1-11}
            \multirow{7}{*}{\rotatebox[origin=c]{90}{\small Finetuning}} & Llama-2-7B             & $38.0 \pm 0.46$                & $26.9 \pm 0.11$ & $53.9 \pm 0.16$                &  & $12.3 \pm 0.18$                & $7.5 \pm 0.06$  & $25.1 \pm 0.25$ &  & $28.1 \pm 1.34$  \\
                                                                         & Llama-2-7B w/ sw       & $39.9 \pm 0.70$                & $25.1 \pm 0.35$ & $51.9 \pm 0.34$                &  & $11.8 \pm 0.14$                & $7.5 \pm 0.12$  & $25.9 \pm 0.41$ &  & $40.1 \pm 1.72$  \\
                                                                         & Llama-2-7B w/ tw       & $41.8 \pm 0.27$                & $23.7 \pm 0.38$ & $51.0 \pm 0.51$                &  & $11.0 \pm 0.30$                & $6.9 \pm 0.20$  & $25.7 \pm 0.45$ &  & $53.3 \pm 2.70$  \\
            \cmidrule{2-11}
                                                                         & Llama-2-13B            & $39.0 \pm 0.66$                & $26.8 \pm 0.38$ & $53.9 \pm 0.34$                &  & $11.8 \pm 0.29$                & $7.3 \pm 0.12$  & $24.8 \pm 0.34$ &  & $33.1 \pm 2.15$  \\
                                                                         & Llama-2-13B w/ sw      & $40.8 \pm 0.36$                & $24.0 \pm 0.44$ & $50.2 \pm 1.06$                &  & $10.8 \pm 0.66$                & $7.2 \pm 0.26$  & $25.6 \pm 0.73$ &  & $51.7 \pm 4.52$  \\
                                                                         & Llama-2-13B w/ tw      & $42.5 \pm 0.31$                & $22.5 \pm 1.47$ & $50.2 \pm 1.59$                &  & $10.7 \pm 0.45$                & $6.8 \pm 0.10$  & $25.8 \pm 0.47$ &  & $60.0 \pm 8.06$  \\
            \cmidrule{2-11}
                                                                         & Llama-2-13B w/ tw + rs & $43.9 \pm 0.33$                & $22.3 \pm 1.32$ & $50.7 \pm 1.55$                &  & $11.8 \pm 0.56$                & $7.8 \pm 0.12$  & $29.5 \pm 0.64$ &  & $62.3 \pm 6.24$  \\
            \bottomrule
        \end{tabular}%
    }
    \caption{Experimental Results. Each metric is reported as $mean \pm std$ across 5 runs. sw = sentence-level loss weights. tw = token-level loss weights. rs = repeated sampling.}
    \label{tab:results}
\end{table*}

\section{Modelling}\label{sec:modelling}
We base our system on Llama 2 7B and 13B  \cite{touvronLlamaOpenFoundation2023} which are pre-trained Transformer models \cite{vaswaniAttentionAllYou2017}. We fine-tune them on the PLABA dataset in a sequence-to-sequence fashion.
For every position $i$ in the target sequence, the cross entropy loss $l$ between the ground truth token $y_i$ and the predicted token $\hat{y}_i$ is backpropagated.

\begin{align*}
    loss = \sum_i l(y_i, \hat{y}_i)
\end{align*}

As illustrated in \autoref{fig:example}, more than half of the tokens from the input are kept in the simplification. Therefore, more than half of the gradients point to the trivial solution of copying the input token.
We argue that this is a suboptimal training setting.
In preliminary experiments, we confirm this by measuring the edit distance between original and simplified sentences, finding that, on average, the model edits less than half as many characters compared to the human annotator.
We hypothesize that editing more would lead to better simplifications, and propose using sentence-level loss weights and token-level loss weights to incentivize the model to edit more.
A higher loss weight $w\in\mathbb{R}^+$ effectively scales the gradient of that sample by $w$.

\subsection{Sentence-level Loss Weights}

We derive the sentence-level loss weights from the edit distance, i.e., a sample with a large distance between the original sentence $x$ and the simplified sentence $y$ is weighted higher and will therefore have a larger impact on the model parameters. More formally, the edit distance $d(\cdot, \cdot)$ is mapped to a loss weight by a function $w(\cdot)$.

\begin{align*}
    loss = w(d(x, y)) \cdot \sum_i l(y_i, \hat{y}_i)
\end{align*}

\begin{figure}[t]
    \includegraphics[width=\linewidth]{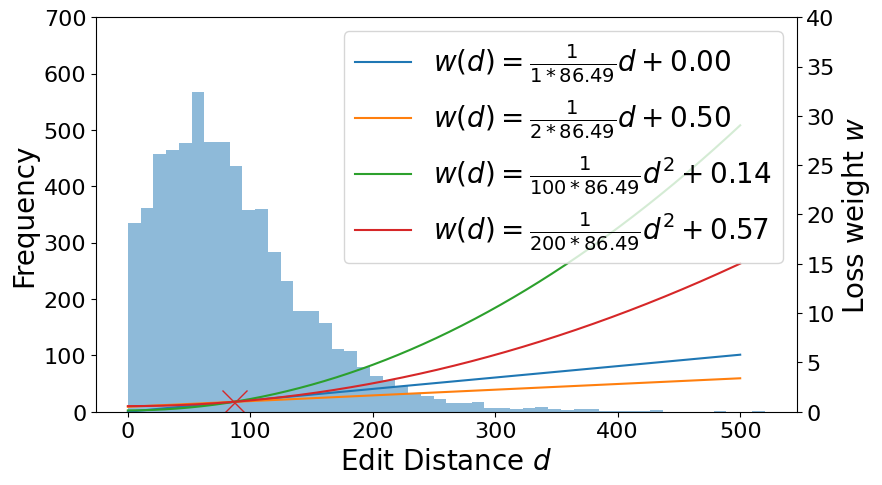}
    \caption{Histogram of edit distances in the dataset and linear and quadratic loss weighting functions $w$. The mean of the distribution is $86.49$, where all loss weighting functions assign a weight of $1$, indicated by the red cross.}
    \label{fig:sweights}
\end{figure}

We experiment with linear and quadratic variants of $w$, as depicted in \autoref{fig:sweights}. For all variants, we define slopes and offsets such that the mean edit distance in the dataset ($86.49$) is assigned a loss weight of $1$.

\subsection{Token-level Loss Weights}

One drawback of sentence-level loss weights is that the losses for all tokens from one sentence are equally re-weighted, including both edited tokens and shared tokens. To get more fine-grained re-weighting of the losses, we further propose token-level loss weights. To that end, we first compute the edit operations required to transform the original sentence $x$
into its simplification $y$ using \texttt{difflib}\footnote{\url{https://docs.python.org/3/library/difflib.html}}. We then increase the loss weight of all target tokens that are part of an edit operation to a hyperparameter $\lambda$.

\begin{align*}
    \lambda_i & = \begin{cases}
        \lambda & \text{if } y_i \text{ is part of an edit operation} \\
        1       & \text{else}
    \end{cases}            \\
    loss      & = \sum_i \lambda_i \cdot l(y_i, \hat{y}_i)
\end{align*}

\subsection{Repeated Sampling}

To further improve the performance in the shared task, we sample from our best model 10 times with varying temperatures. We then prompt ChatGPT with the generated simplifications and instruct it to select the best one in a few-shot setting. 

\section{Results}\label{sec:results}

In this section, we analyze the results of our empirical evaluation shown in \Cref{tab:results}.

\subsection{Evaluation and Experimental Setup}

In the shared task, the simplifications are automatically evaluated using the SARI metric \cite{xuOptimizingStatisticalMachine2016}, measuring the $F_1$ score of n-gram edit operations. The shared task submissions are furthermore manually evaluated based on sentence simplicity, term simplicity, term accuracy, fluency, completeness, and faithfulness. In addition to the shared task evaluation on the blind test set, we conduct a comprehensive automatic evaluation of all models on the labelled test set. To measure the similarity to the ground truth, we use SARI, BLEU \cite{papineniBleuMethodAutomatic2002} and ROUGE-L \cite{linROUGEPackageAutomatic2004}. We furthermore use Flesch-Kincaid Grade Level (FKGL, \citealp{flesch1943marks}), the number of difficult words\footnote{\url{https://github.com/textstat/textstat}} and the total number of words to measure the simplicity of the generated outputs.
Moreover, we report the edit distance \cite{levenshtein1966binary} to measure how many characters were edited by the models. 

For all model configurations, we train and generate five times in order to approximate statistical significance using standard deviations. To save compute, we perform all fine-tuning using LoRA \cite{huLoRALowRankAdaptation2021a}, i.e., only low-rank decompositions of the full parameter matrices are trained. The hyperparameters are reported in \Cref{sec:hyperparameters}.

\subsection{SARI and Edit Distance} 

\begin{figure}[t]
    \centering
    \includegraphics[height=5cm]{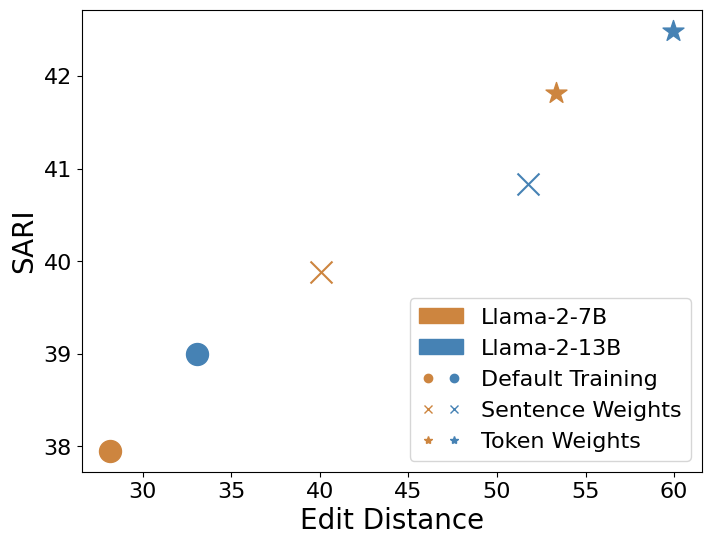}
    \caption{Scatter plot of edit distances and SARI scores for 7B and 13B models fine-tuned without loss weights, with sentence-level loss weights, and with token-level loss weights. The Pearson correlation is $0.99$.}
    \label{fig:corr}
\end{figure}

Both sentence-level and token-level loss weights lead to much higher edit distances compared to training without loss weights.
In particular, the edit distance increases by 81\% with token-level and 56\% with sentence-level loss weights.
We furthermore observe a very strong correlation ($0.99$) between edit distance and SARI as depicted in \autoref{fig:corr}, confirming our hypothesis that incentivizing the model to edit more will also lead to better simplifications.
Adding token-level loss weights in the 7B model is much more effective ($+3.9$ SARI) than scaling up to 13B without loss weights ($+1.1$ SARI) while being significantly cheaper during training and inference.

\subsection{$\lambda$ as Control Parameter}

\begin{figure}[t]
    \centering
    \includegraphics[height=5cm]{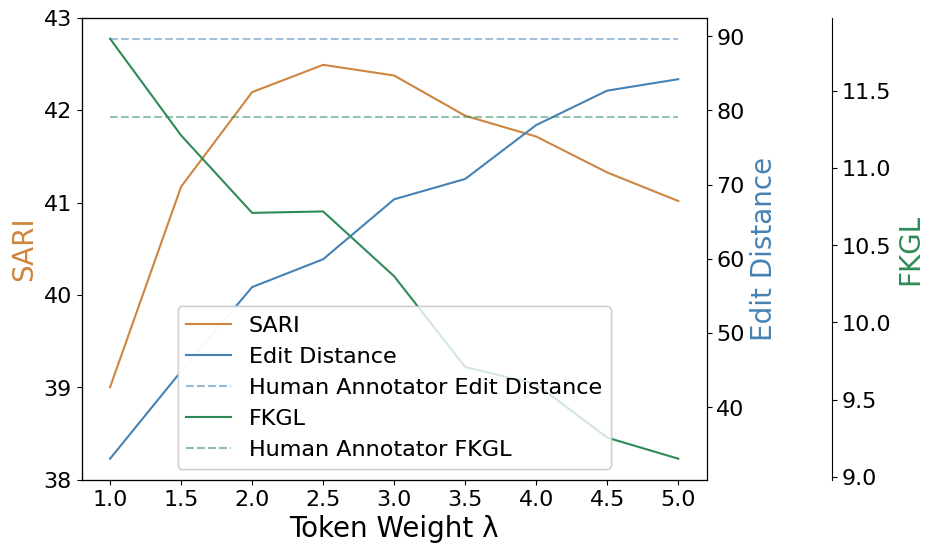}
    \caption{SARI score and edit distance for different values of $\lambda$ in token-level loss weights. Increasing $\lambda$ monotonically leads to higher edit distances and lower FKGL, whereas SARI peaks at $\lambda=2.5$.}
    \label{fig:lambda}
\end{figure}

Motivated by the finding that incentivizing the model to edit more also leads to higher SARI scores, we experiment with the hyperparameter $\lambda$ which determines the weight of edited tokens during training. The results are depicted in \autoref{fig:lambda}. Increasing $\lambda$ monotonically leads to a higher edit distance and lower FKGL.
That is, $\lambda$ can be effectively used as a control mechanism for the edit distance and FKGL.
Contrary to the edit distance, SARI does not monotonically increase with $\lambda$ but peaks at $\lambda=2.5$. Interestingly, this peak occurs far before the model's edit distance has matched that of the human annotators ($60.0$ vs $89.7$).
There are two possible reasons for this: either the model starts performing incorrect edits, or the edits are correct but not present in the single reference simplification.  

\subsection{Repeated Sampling}

Repeated sampling further improves the SARI score by almost $1.5$ percentage points. On the other hand, the simplifications are more complex compared to sampling just a single time, indicated by higher FKGL, more difficult words and longer sentences. I.e., ChatGPT seems to be biased towards the more complex simplifications. We leave experimentation with alternative critics to future work.

\begin{table*}[t!]
    \resizebox{\textwidth}{!}{%
        \footnotesize
        \begin{tabular}{ll}
            \toprule
            Author & Text \\ \midrule

            Original & 
            \parbox{0.8\linewidth}{The mechanisms and kinetics of the immunosuppressive effects of alternate-day prednisone were investigated in a group of patients with a variety of inflammatory diseases receiving a range of alternate-day prednisone doses from 5 to 120 mg.} \vspace{1em}\\ 

            Human & 
            \parbox{0.8\linewidth}{The process and reactions of the immune system from taking alternate- or every-other-day prednisone (a type of steroid drug that works by lowering the activity of the immune system) are investigated in a group of patients with different inflammatory diseases (diseases in which the immune system attacks the body's own tissues).} \vspace{1em} \\

            Model & 
            \parbox{0.8\linewidth}{We looked at how the immune system was affected by alternate-day prednisone. We looked at this in a group of patients with different inflammatory diseases. The patients were taking different doses of alternate-day prednisone.} \\

            \cmidrule{1-2}
            Original & 
            \parbox{0.8\linewidth}{The first trimester screening programme offers a noninvasive option for the early detection of aneuploidy pregnancies.} \vspace{1em}\\ 

            Human & 
            \parbox{0.8\linewidth}{The first trimester screening program offers a noninvasive (nonsurgical) option for the early detection of aneuploidy pregnancies. Aneuploidy refers to the conditions of having an abnormal number of chromosomes.} \vspace{1em} \\

            Model & 
            \parbox{0.8\linewidth}{ The first trimester screening program offers a noninvasive option for the early detection of aneuploidy pregnancies.} \\

            \bottomrule
        \end{tabular}
    }
    \caption{Examples of original sentence, human-written simplification and model-written simplification for qualitative analysis. The model is trained with token-level loss weights and $\lambda=2.5$.}
    \label{tab:examples}
\end{table*}

\subsection{Zero-Shot}

Fine-tuning outperforms zero-shot prediction by a large margin across models in our experiments.
We find that Llama-2 is unable to generate only the simplified sentence, but also includes conversational phrases like \enquote{Sure, here you go,}
as well as additional explanations.
We experimented with different instructions to avoid these pitfalls, but found that they are not effective enough to consistently control the target format. We leave experimentation with few-shot approaches to future work.

\subsection{BLEU and ROUGE}

In our experiments, BLEU and ROUGE scores seem to be negatively correlated with SARI. Not performing any edits (copy baseline) actually achieves the highest BLEU and ROUGE scores. This could be because many of the tokens are not edited by the human annotators, while there are many possible variants of good simplification operations. That is, editing a token will likely decrease n-gram precision even if the edit is good. We see similar trends in prior work on simplification \cite{attalDatasetPlainLanguage2023,liLargeLanguageModels2023}. This underlines the importance of simplification-specific evaluation metrics that take the original sentence into account, as well as using multiple reference simplifications.

\subsection{Human Evaluation}

The human evaluation was performed by the shared task organizers. Annotators were asked to rate samples on multiple criteria, where each criterion could be rated as -1, 0 or 1. The final score per criterion was then calculated by summing all values and mapping the interval to a scale from 0 to 100. We submitted the predictions from Llama 2 13B trained with token-level loss weights and $\lambda=2.5$, i.e., the model with the best SARI score in our evaluation. Overall, we were first in the accuracy category including the completeness and faithfulness criteria with a mean score of 94.87. The model therefore generated simplifications that matched the input very well in terms of information content. In the simplicity category including sentence simplicity, term simplicity, term accuracy and fluency, we ranked second with a mean score of 88.86. Among these criteria, all ratings were above 90, except for term simplicity with a value of 77.25. That means that the model did not always explain or replace complex terms.

\section{Discussion}\label{sec:discussion}
\subsection{Low FKGL with Few Edits}

As shown in \autoref{tab:results}, the fine-tuned models typically achieve a lower FKGL score, despite editing much fewer characters on average. One might conclude that the model is able to simplify the sentence more efficiently than the human. 
We manually analyze cases where the model has much lower FKGL and edit distance and find that this is rather due to different simplification styles. 
An example is depicted in the first row of \autoref{tab:examples}. 
The human annotator added long term explanations in parentheses, whereas the model split up the sentence into 3 shorter sentences. 
On the other hand, the model did not explain or simplify the term \enquote{prednisone.} Generally, we observe that the model, compared to the human annotator, focuses more on grammatical simplification than term simplification. 
This insight is also reflected in the human evaluation, where our model achieved much higher sentence simplicity than term simplicity.

\subsection{Controllability}

When varying the $\lambda$ hyperparameter for the token-level loss weights, we found that $\lambda$ strongly correlates with edit distance and FKGL.
We conclude that the method has potential applications for controllable generation. However, unlike many other control mechanisms, such as control tokens \cite{martinControllableSentenceSimplification2020}, $\lambda$ has to be set during training. Control at inference time would therefore require multiple training runs. Nevertheless, $\lambda$ can be used to tune the model towards a target simplicity level during training.

\subsection{Generalization to Other Domains}

The proposed loss weights have shown significant improvements on the PLABA dataset. Yet, the method should be evaluated on additional datasets to gain more broadly applicable insights. We argue that most sentence pairs in simplification datasets share a large portion of the tokens. Therefore, improvements through loss weights are plausible across domains.

\subsection{Validity of Automated Metrics}

We used several automated metrics in our empirical evaluation. BLEU and ROUGE do not seem to be informative, as copying the input achieves very competitive scores. That is, shared tokens pose challenges not only during training but also during evaluation. SARI has been shown to correlate with human judgement, but we find patterns where SARI produces unintuitive results. SARI often does not reflect that terms have been properly substituted or explained. For instance, the model-generated output in line 2 of \autoref{tab:examples} achieves a very high SARI score of $72.4$ despite not explaining \enquote{noninvasive} and \enquote{aneuploidy.} This underlines the importance of task-specific metrics that align with the human quality criteria. For instance, term simplicity could be indicated by the number of difficult terms that are either substituted or explained in parentheses, measured by semantic similarity.

\section{Conclusion}\label{sec:conclusion}
In this work, we first identified the issue that tokens that are not edited by the human annotator have gradients pointing to the trivial solution of copying the input. 
We indeed found that models trained with equally weighted gradients tend to edit very conservatively. 
We hypothesized that incentivizing the model to edit more will also lead to better simplifications, as measured by SARI. 
We proposed token-level and sentence-level loss weights based on the edit operations and the edit distance, respectively. 
Both led to significantly increased edit distance and decreased FKGL, while also improving SARI. 
Due to the more fine-grained adjustment, token-level loss weights performed best and led to an increase of $3.9$ percentage points in SARI compared to regular training.
We additionally used repeated sampling with ChatGPT as critic and ranked first in the automatic evaluation with a gap of $2.5$ percentage points in SARI to second place.

\bibliographystyle{acl_natbib}
\bibliography{custom}

\appendix

\section{Hyperparameters}\label{sec:hyperparameters}

\FloatBarrier
\begin{center}
    \begin{tabular}{lll}
        \toprule
        Category                    & Parameter      & Value  \\
        \midrule
        \multirow{6}{*}{Training}   & Learning Rate  & $5e-5$ \\
                                    & Batch Size     & 32     \\
                                    & Epochs         & 3      \\
                                    & Precision      & BF16   \\
                                    & LoRa Rank      & 8      \\
                                    & LoRa $\alpha$  & 32     \\
        \cmidrule{1-3}
        \multirow{3}{*}{Generation} & Temperature    & 0.6    \\
                                    & Top-P          & 0.7    \\
                                    & Max New Tokens & 128    \\
        \bottomrule
    \end{tabular}
\end{center}
\FloatBarrier

\section{Computational Efforts}

All computations were performed using A100 GPUs with 80GB of VRAM. 
Every run was executed on a single GPU and took about 2 hours for training and generation.
We estimate the total computational efforts to about 200 GPU-hours. 

\end{document}